%% file: main.tex
\documentclass{article}
\usepackage{spconf,amsmath,graphicx}
\usepackage{subfig}
\usepackage{tabularx}
\usepackage{graphicx}
\usepackage{tabularx}
\usepackage{amssymb}
\usepackage{algorithm}
\usepackage{multirow}
\usepackage{subfig}
\usepackage{graphics}
\usepackage{amsmath}
\usepackage{amsfonts}
\usepackage{epstopdf}
\usepackage{booktabs}
\usepackage{url}
\usepackage{color}
\usepackage{xspace}
\usepackage{balance}
\usepackage{cleveref}

\crefformat{footnote}{#2\footnotemark[#1]#3}

\def\x{{\mathbf x}}

\def\@onedot{\ifx\@let@token.\else.\null\fi\xspace}

\makeatother

\newcommand{\UBM}{\boldsymbol{m}}
\newcommand{\U}{\boldsymbol{U}}
\newcommand{\D}{\boldsymbol{D}}
\newcommand{\N}{\boldsymbol{N}}
\newcommand{\image}{\boldsymbol{I}}
\newcommand{\observations}{\boldsymbol{O}}
\newcommand{\obs}{\boldsymbol{o}}

\newcommand{\normal}{\mathcal{N}}
\newcommand{\var}{\boldsymbol{\Sigma}}
\newcommand{\client}{\boldsymbol{s}}
\newcommand{\z}{\boldsymbol{z}}
\newcommand{\f}{\boldsymbol{f}}
\newcommand{\isvmean}{\boldsymbol{u}}

\title{Modelling Local Deep Convolutional Neural Network Features\\to Improve Fine-Grained Image Classification}
\name{ZongYuan Ge, Chris McCool, Conrad Sanderson, Peter Corke}
\address
  {
  Queensland University of Technology, Brisbane, QLD 4000, Australia\\
  NICTA, PO Box 10522, Adelaide St, Brisbane, QLD 4001, Australia\\
  }

\begin{document}
\maketitle

\input{abstract}
\input{introduction}

\input{prior_work}

\input{proposed_method}

\input{experiment}
\input{conclusion}

\balance
\small

\bibliographystyle{ieee}
\bibliography{refs}

\end{document}

%% file: abstract.tex
\begin{abstract}
\vspace{-0.5ex}
We propose a local modelling approach using deep convolutional neural networks (CNNs) for fine-grained image classification. 
Recently, deep CNNs trained from large datasets have considerably improved the performance of object recognition.
However, to date there has been limited work using these deep CNNs as local feature extractors.
This partly stems from CNNs having internal representations which are high dimensional,
thereby making such representations difficult to model using stochastic models.
To overcome this issue, we propose to reduce the dimensionality of one of the internal fully connected layers,
in conjunction with layer-restricted retraining to avoid retraining the entire network.
The distribution of low-dimensional features obtained from the modified layer is then modelled using a Gaussian mixture model.
Comparative experiments show that considerable performance improvements can be achieved on the challenging Fish and UEC FOOD-100 datasets.
\end{abstract}
\begin{keywords}
fine-grained classification, deep convolutional neural networks, session variation modelling, Gaussian mixture models.
\end{keywords}

%% file: introduction.tex
\vspace{-2ex}
\section{Introduction}
\vspace{-1ex}
\label{sec:intro}

\begin{figure}[!tb]%
  \centering
  \begin{minipage}{0.85\columnwidth}
    \centering
  
    \begin{minipage}{1\textwidth}
      \begin{minipage}{0.49\textwidth}
        \centering
        \includegraphics[width=1\textwidth]{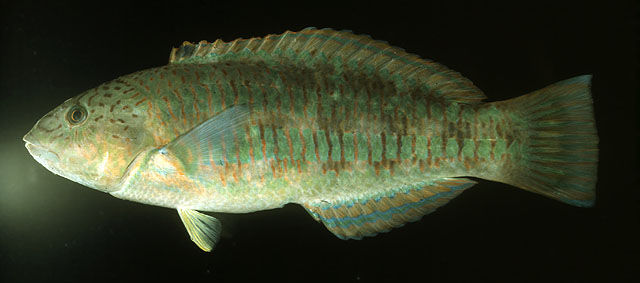}
      \end{minipage}
      \hfill
      \begin{minipage}{0.49\textwidth}
        \centering
        \includegraphics[width=1\textwidth]{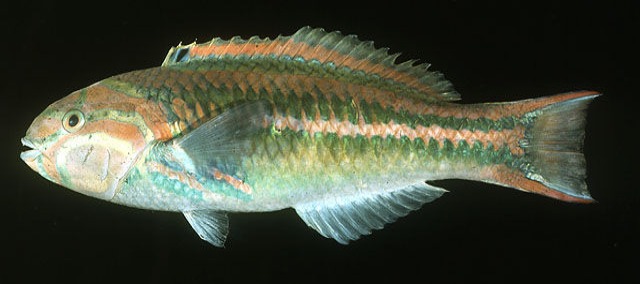}
      \end{minipage}
    \end{minipage}
  
    \begin{minipage}{1\textwidth}
      \begin{minipage}{0.49\textwidth}
        \centering
        {\footnotesize Thalassoma Trilobatum}
      \end{minipage}
      \hfill
      \begin{minipage}{0.49\textwidth}
        \centering
        {\footnotesize Thalassoma Quinquevittatum}
      \end{minipage}
    \end{minipage}
    
    \begin{minipage}{1\textwidth}
      \begin{minipage}{0.49\textwidth}
      \end{minipage}
      \hfill
      \begin{minipage}{0.49\textwidth}
      \end{minipage}
    \end{minipage}

    \begin{minipage}{1\textwidth}
      \begin{minipage}{0.49\textwidth}
        \centering
        \includegraphics[width=1\textwidth]{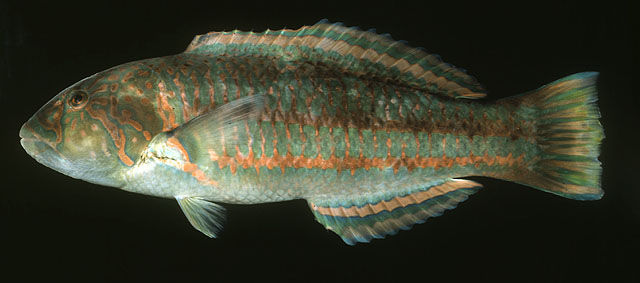}
      \end{minipage}
      \hfill
      \begin{minipage}{0.49\textwidth}
        \centering
        \includegraphics[width=1\textwidth]{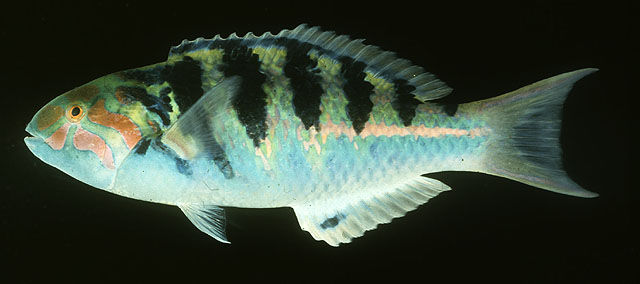}
      \end{minipage}
    \end{minipage}

    \begin{minipage}{1\textwidth}
      \begin{minipage}{0.49\textwidth}
        \centering
        {\footnotesize Thalassoma Purporeum}
      \end{minipage}
      \hfill
      \begin{minipage}{0.49\textwidth}
        \centering
        {\footnotesize Thalassoma Hardwicke}
      \end{minipage}
    \end{minipage}
    
  ~
  
    \begin{minipage}{1\textwidth}
      \begin{minipage}{0.49\textwidth}
        \centering
        \includegraphics[width=0.48\textwidth]{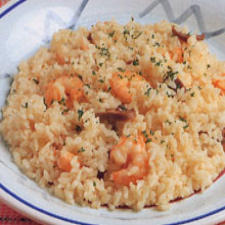}
        \hfill
        \includegraphics[width=0.48\textwidth]{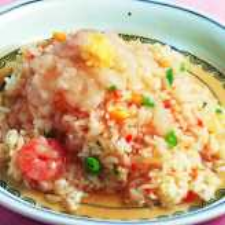}
      \end{minipage}
      \hfill
      \begin{minipage}{0.49\textwidth}
        \centering
        \includegraphics[width=0.48\textwidth]{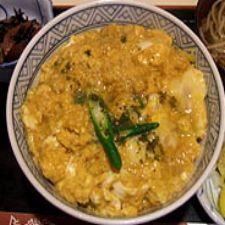}
        \hfill
        \includegraphics[width=0.48\textwidth]{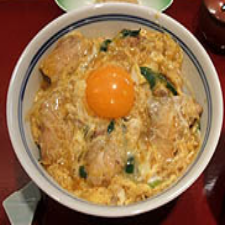}
      \end{minipage}
    \end{minipage}
  
    \begin{minipage}{1\textwidth}
      \begin{minipage}{0.49\textwidth}
        \centering
        {\footnotesize Fried Rice}
      \end{minipage}
      \hfill
      \begin{minipage}{0.49\textwidth}
        \centering
        {\footnotesize Chicken Rice}
      \end{minipage}
    \end{minipage}
    
    \begin{minipage}{1\textwidth}
      \begin{minipage}{0.49\textwidth}
      \end{minipage}
      \hfill
      \begin{minipage}{0.49\textwidth}
      \end{minipage}
    \end{minipage}

    \begin{minipage}{1\textwidth}
      \begin{minipage}{0.49\textwidth}
        \centering
        \includegraphics[width=0.48\textwidth]{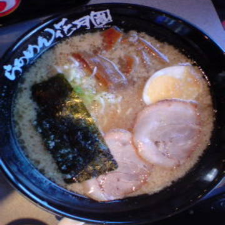}
        \hfill
        \includegraphics[width=0.48\textwidth]{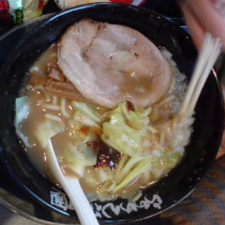}
      \end{minipage}
      \hfill
      \begin{minipage}{0.49\textwidth}
        \centering
        \includegraphics[width=0.48\textwidth]{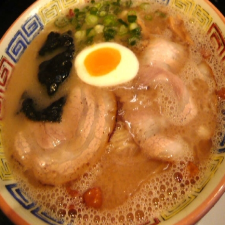}
        \hfill
        \includegraphics[width=0.48\textwidth]{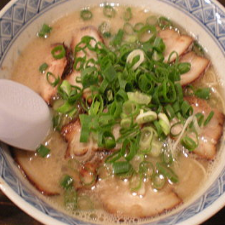}
      \end{minipage}
    \end{minipage}

    \begin{minipage}{1\textwidth}
      \begin{minipage}{0.49\textwidth}
        \centering
        {\footnotesize Ramen}
      \end{minipage}
      \hfill
      \begin{minipage}{0.49\textwidth}
        \centering
        {\footnotesize Beef Noodle}
      \end{minipage}
    \end{minipage}

  \end{minipage}
  
  \vspace{-0.5ex}
  \caption
    {
    \small
    First two rows show example images of four fish species,\
    which have low inter-class variation: similar visual appearance despite being distinct species. (Images taken by J.E.~Randall).
    The last two rows show images of four food dishes, with each dish type having high intra-class variation.
    }%
  \label{fig:similar_fish_food}
  \hrule
  \vspace{-0.5cm}
\end{figure}

Fine-grained image classification refers to the task of recognising the class or subcategory (for instance the particular fish species) under the same basic category such as bird or fish species~\cite{anantharajah2014local,zhang2014part}.  
This is a challenging task for two reasons.
First, some classes (species) from the same category, such as fish, can appear to be very similar in terms of appearance leading to low inter-class variation.
Second, there is a high degree of variability in the instances of the same classes due to environmental and illumination variations leading to high intra-class variation.
Fig.~\ref{fig:similar_fish_food} shows examples of both issues.

An approach to tackling these two issues is to extract local region descriptors and to model them.
Such an approach has previously been popular for recognition of faces~\cite{Lucey2004, Sanderson2003} and fish~\cite{anantharajah2014local}. 
These approaches typically divide the image into patches (or blocks),
with each patch considered to be an independent (and partial) observation of the object.
Each patch is then represented by a feature vector and the distribution of all of these features vectors, from an image, is then modelled using a Gaussian mixture model (GMM).
The feature vector to represent each patch has usually been obtained from a transform such as the 2D discrete cosine transform~\cite{Sanderson2003}.

Recently, feature learning through the use of deep convolutional neural networks (CNNs) has led to considerable improvements for object recognition~\cite{krizhevsky2012imagenet}.
These deep CNN feature representations are trained on large datasets such as ImageNet~\cite{deng2009imagenet} which has {\small $1,000$} general object categories.
It has been shown that these learnt features can be used to obtain impressive results for other recognition tasks when used as a global image representation~\cite{razavian2014cnn}.
However, to the best of our knowledge no work has examined how to use these learnt features as a local feature extractor for use with well known statistical modelling approaches such as GMMs.

To use these deep CNN features as a local feature extractor two issues need to be addressed.
First, deep CNNs such as~\cite{krizhevsky2012imagenet} generally have an internal representation which is high dimensional, leading to the curse of dimensionality~\cite{Bishop06} for local modelling techniques such as GMMs.
Second, we need to develop an efficient and effective method to retrain a deep CNN containing millions of weights using a relatively small set of images specific to a fine-grained class.
In this paper we address both of these issues.

Inspired by recent work that has shown how to optimise deep CNN features for small datasets using fine-tuning~\cite{zhang2014part}, we propose a method to obtain a low-dimensional deep CNN representation that can be used as a local feature descriptor.
Specifically, we propose to explicitly reduce the dimensionality of one of the internal fully connected layers,
in conjunction with using layer-restricted retraining to avoid retraining the entire network.
We demonstrate empirically that the proposed approach leads to considerable performance improvements for two fine-grained image classification tasks:
fish recognition~\cite{anantharajah2014local} and food recognition~\cite{matsuda12}.

We continue the paper as follows.
In Section~\ref{sec:prior_work} we briefly describe the image classification approach
based on statistical modelling of local features and inter-session variability modelling.
The approach is used as a base upon which we build on in Section~\ref{sec:proposed},
where we learn a low-dimensional deep CNN representation that can be used as local feature descriptor.
Comparative experiments are given in Section~\ref{sec:exp},
followed by the main findings and future directions in Section~\ref{sec:conclusion}.

%% file: prior_work.tex
\vspace{-1.5ex}
\section{Modelling Local Image Features}
\label{sec:prior_work}
\vspace{-1ex}

Modelling the distribution of local features has been explored by several researchers~\cite{Lucey2004, Sanderson2003, McCool2013}.
In general, these methods divide the $j$-th image of the $i$-th class, $\image_{i,j}$, into $N$ overlapping patches.
Each patch is represented by an $M$-dimensional feature vector, of low dimensionality, to yield the set of $N$ feature vectors $\observations_{i, j} = \left[\obs_{i, j, 1}, \dots, \obs_{i, j, N}\right]$. 
The distribution of the vectors is then modelled using a GMM to obtain a prior model, referred to as a universal background model (UBM), that represents the basic category in question (eg.~fish, food).

This UBM representation forms the basis which many feature modelling methods use.
It can be used as a probabilistic bag-of-words representation~\cite{Sanderson09} or a model can be derived for each class by performing mean-only relevance MAP adaptation~\cite{Lucey2004}.
Another extension is to perform inter-session variability (ISV) modelling~\cite{McCool2013} which learns those variations that can make one instance (image) of the same class look different to another image of the same class.

Irrespective of the specific method they all rely on a GMM which is known to perform poorly for high-dimensional data~\cite{Bouveyron06highdimensional}.
This is partly due to the curse of dimensionality where it becomes difficult to estimate a large number of parameters when there is limited data.
To avoid this we will show how to learn a low-dimensional deep CNN representation, however, before proceeding to this we first describe the GMM feature modelling methods that we use in this work.

\vspace{-2ex}
\subsection{GMM Feature Modelling}
\vspace{-1ex}

We use two feature modelling approaches in this work, GMM mean-only MAP adaptation and its extension ISV.
These two are chosen as they have been shown to provide consistently good performance~\cite{McCool2013}.

GMM mean-only MAP adaptation takes the prior model (UBM) and adapts just the means using the enrollment data of the $i$-th class $\observations_{i}$; all of the features for the $J_{i}$ enrollment images.
Using supervector notation~\cite{McCool2013}, this is written as
\vspace{-1ex}
\begin{equation}
\label{eq:prior_map}
  \client_{i} = \UBM + \D \z_{i}, 
\vspace{-1ex}
\end{equation}
\noindent
where $\client_{i}$ is the mean supervector for the $i$-th class, $\UBM$ is the mean supervector of the UBM (the prior), $\z_{i}$ is a normally distributed latent variable, and $\D$ is a diagonal matrix that incorporates the relevance factor and the covariance matrix and ensures the result is equivalent to mean-only relevance MAP adaptation.

ISV is an extension of the GMM mean-only MAP model which learns a sub-space which models and suppresses session variation~\cite{McCool2013}.
It includes a subspace $\U$ to cope with session variation and is written in supervector notation as
\vspace{-1ex}
\begin{equation}
\label{eq:prior_isvmean}
  \isvmean_{i,j} = \UBM + \U \x_{i,j} + \D \z_{i},
\vspace{-1ex}
\end{equation}
\noindent where $\x_{i,j}$ is the latent session variable and is assumed to be normally distributed.
Suppressing the session variation is done by jointly estimating the latent variables $\z_{i}$ and $\left[\x_{i,1}, \dots \x_{i,J_{i}}\right]$
followed by discarding the latent session variables to give
\vspace{-1ex}
\begin{equation}
\label{eq:clean_ISV}
  \client_{ISV,i} = \UBM + \D \z_{i}, 
\vspace{-1ex}
\end{equation}

For both of these methods, the log-likelihood ratio is used to determine if the $t$-th test image $\image_t$ was most likely produced by class $i$.
This is efficiently calculated using the linear scoring approximation~\cite{Glembek2009} which for GMM mean-only MAP~is
\vspace{-1ex}
\begin{equation}
  h_{linear}\left(\observations_{t}, \client_{i}\right) = \left(\client_{i} - \UBM \right)^{T} \var^{-1} \f_{t \mid \UBM},
\end{equation}
\noindent
and for ISV it is
\vspace{-1ex}
\begin{equation*}
  h_{ISV}\left(\observations_{t}, \client_{i}\right)
  \mbox{=}
  \left(\client_{ISV,i} \mbox{~-~} \UBM \right)^{T} \var^{\mbox{-}1} \left( \f_{t \mid \UBM} \mbox{~-~} \N_{t} \U \x_{t \mid \UBM} \right),
\vspace{-1ex}
\end{equation*}
\noindent
where the diagonal matrix $\var$ is formed by concatenating the diagonals of the UBM covariance matrices,
$\f_{t \mid \UBM}$ is the supervector of mean normalised first order statistics,
and $\N_{t}$ contains the zeroth order statistics for the test sample in a block diagonal matrix~\cite{McCool2013}.

%% file: proposed_method.tex
\vspace{-2ex}
\section{Proposed Method}
\vspace{-1ex}
\label{sec:proposed}

\begin{figure}[!t]%
  \centering
    \includegraphics[width=1\columnwidth]{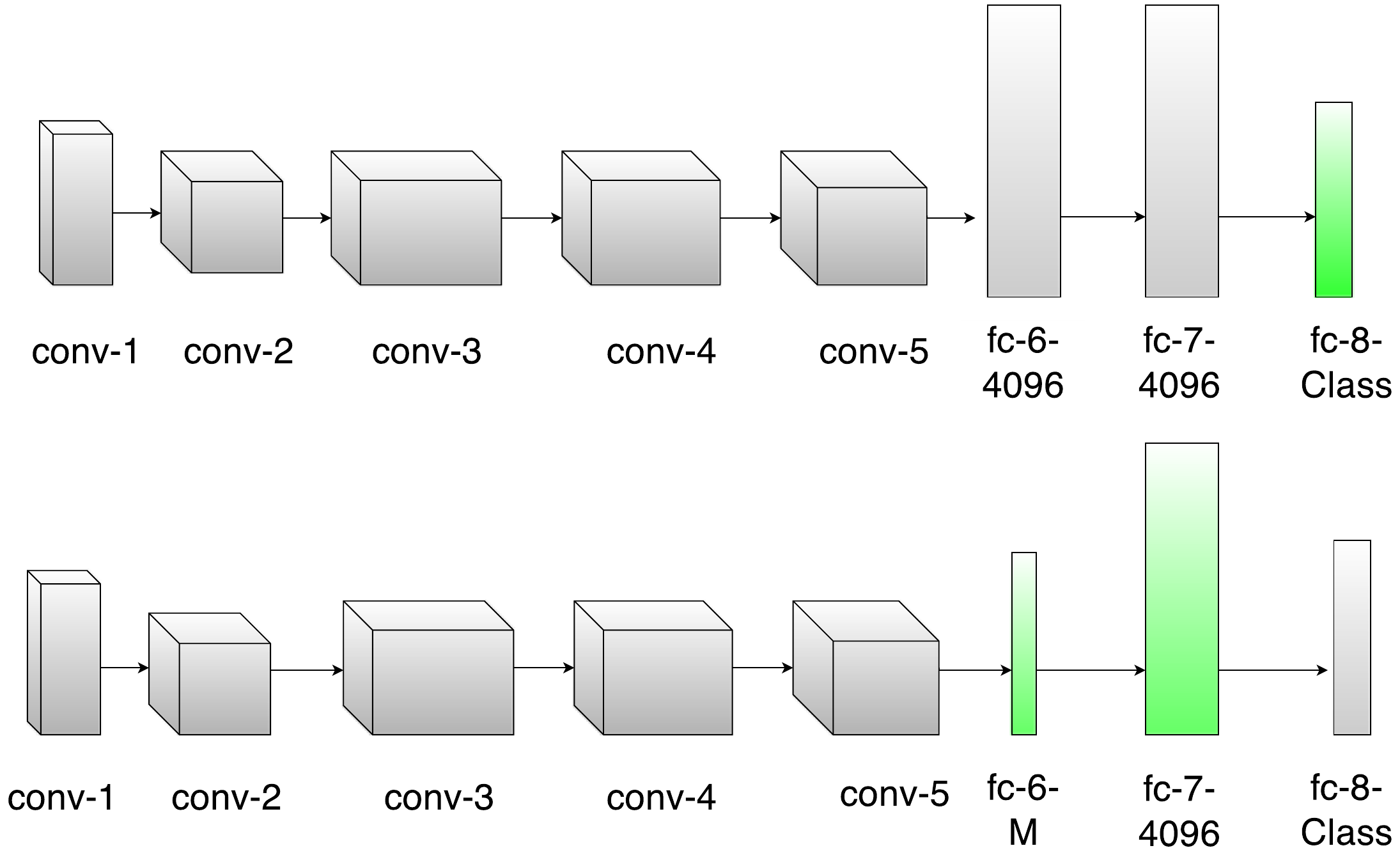} 
  \vspace{-0.5cm}
  \caption
    {
    \small
    Modifying and retraining the deep CNN through a 2 step procedure.
    For each step we have shaded in green the parts of the network that are changed and retrained.
    First step: the highlighted \mbox{fc-8} layer is modified to have only as many outputs as the number of dataset specific classes.
    The layer is retrained, while all the other parameters remain fixed.
    Second step: the highlighted \mbox{fc-6} layer is changed to map to only $M$ outputs,
    followed by training the \mbox{fc-6} layer in conjunction with the highlighted \mbox{fc-7} layer,
    while keeping the remaining parameters fixed.
    The output of the fc-6 layer is used as a local feature extractor.
    }%
  \label{fig:cnn}
  \hrule
  \vspace{-1ex}
\end{figure}

To extract features from local patches, we aim to learn a low-dimensional deep CNN representation which we refer to as a low-dimensional CNN feature vector (LDCNN).
This is in contrast to the high dimensional representation ($4096$ dimensions) that is usually obtained from the fully connected layer \mbox{(fc-6)} of the pretrained deep CNN~\cite{krizhevsky2012imagenet}, the structure of this network can be seen in Fig.~\ref{fig:cnn}.
Such high dimensional representations are difficult to be effectively modeled with a stochastic model such as a GMM, as such we aim to learn a low-dimensional representation (LDCNN) whose dimensionality $M$ is much less than $4096$.
To reduce the dimensionality while preventing the parameters from overfitting in the large CNN architecture, we propose a two step modification for the network.




In the first step, using the pretrained network of~\cite{krizhevsky2012imagenet} as a starting point, we modify the final output layer (fc-8) to have outputs for the {\small $N_c$} training classes.
The weights are randomly initialised\footnote{\label{foot:random} Random initialisation is performed by drawing from $\normal \left( 0, 0.01^2 \right)$.} and retraining is then conducted such that only the fc-8 layer is updated using a learning rate of $0.01$.
This process equates to a multiclass linear regression, using the pretrained network as a feature extractor.
It converges after a few thousand iterations.  

In the second step we replace the two fully connected layers fc-6 and fc-7 and retrain only these two layers with the other layers fixed.
We replace the original {\small $4096$} dimension fc-6 layer with a new {\small $M$}-dimensional fc-6 layer that is randomly initialised\cref{foot:random}, where {\small $M \ll 4096$}.
Features extracted from this layer are referred to as LDCNN.
The fc-7 layer is also replaced and randomly initialised\cref{foot:random} as fc-6 and fc-7 are densely connected.
However, when we retrain the network, fc-7 retains its original dimensionality of {\small $4096$}.
Retraining is then performed using back propagation and stochastic gradient descent to update only these two layers.
The learning rate is initially set to {\small $0.01$} but this rate reduces by a factor of {\small $10$} for every {\small $1000$} iterations throughout training process.
In this way, all pretrained convolutional layer filters from the original network~\cite{krizhevsky2012imagenet} are retained.

%
%

%% file: experiment.tex
\vspace{-2ex}
\section{Experiments}
\label{sec:exp}
\vspace{-1.5ex}



We evaluate our approach on two fine-grained image datasets: Fish~\cite{anantharajah2014local} and UEC FOOD-100~\cite{matsuda12}.
For both datasets we present two baseline systems, both of which perform classification using an SVM and extract a single global CNN feature to represent each image.
The first baseline extracts a single global feature vector using fc-6 of the pre-trained deep CNN~\cite{krizhevsky2012imagenet} ({\small $4096$}~dimensions); we refer to this as \textbf{SVM-CNN}.
The second baseline extracts a single global feature vector using the re-trained low-dimensional CNN feature (LDCNN) vector; we refer to this as \textbf{SVM-LDCNN}.

The local features modelling results (GMM), where the image is divided into $N$ overlapping patches, use two feature extractors.
These feature extractors obtain an $M$-dimensional feature vector from each of the $N$ patches which is then modelled using a GMM.
The first, \textbf{GMM-LDCNN}, uses the proposed low-dimensional CNN feature vector (LDCNN) to obtain the $M$-dimensional feature vector.
The second, \textbf{GMM-PCA-CNN}, uses fc-6 pre-trained deep CNN~\cite{krizhevsky2012imagenet} ({\small $4096$}~dimensions) and learns a transform using principal component analysis (PCA)~\cite{Fukunaga90} to reduce the dimensionality to {\small $M$}.

When we perform local feature modelling (GMM) a range of parameters are varied.
The number of components evaluated for the GMM were \mbox{\small $C$ = $\left[128, 256, 512, 1024\right]$},
the size of the ISV subspace was \mbox{\small $N_{U}$ = $\left[2, 4, 8, \dots, 256 \right]$},
and the range of block sizes \mbox{\small $B$ = $\left[32, 64, 96, 128\right]$}.
For both datasets the images were resized to be {\small $256 \times 256$}.
Caffe~\cite{jia2014caffe} was used to extract and retrain the CNN features and Bob~\cite{bob2012} was used to learn the GMM and ISV models.

\vspace{-1ex}
\subsection{Fine-Grained Fish Classification}
\vspace{-1ex}

We use the Fish image dataset from~\cite{anantharajah2014local} which consists of {\small $3,960$} images collected from {\small $468$} species.
This dataset contains images captured in different conditions, defined as ``controlled'', ``out-of-the-water'' and ``in-situ''. 
The ``controlled'' images consist of fish specimens with controlled background and illumination. The ``in-situ'' images are underwater images of fish in their natural habitat and the ``out-of-the-water'' images consist of fish specimens taken out of the water with a varying background.

Following the defined protocols, the dataset is split into three sets: a training set (\textit{train}) to learn/derive UBM GMM models; a development set (\textit{dev}) to determine the optimal parameters and decision threshold for our models and an evaluation set (\textit{eval}) to measure the final system performance.
There are two protocols: protocol 1a evaluates the system performance when high quality (``controlled'') data is used to enrol classes and protocol 1b evaluates the system performance when low quality (``in-situ'') data is used to enrol classes.
For both protocols, the same test imagery (a mix of ``controlled'', ``in-situ'' and ``out-of-the-water'' images) is used.
The local modelling approach used for these experiments was the ISV extension of the GMM approach as this provided a considerable boost for the initial experiments; we refer to this as \textbf{GMM-LDCNN}.

It has been shown in~\cite{anantharajah2014local} that incorporating spatial information can be advantageous, and as such we further propose to extend the GMM-LDCNN approach
by adding the spatial location {\small $(x, y)$} to each local feature vector prior to modelling; we refer to this method as GMM-LDCNN-xy.

The results in Table~\ref{tbl:fish_svm} show that 
in contrast to global features, local modelling provides notable improvements:
the two baseline systems (SVM-CNN and SVM-LDCNN) which use global features perform worse than the previous state-of-the-art local ISV modelling approach (Local~GMM).
Furthermore, our local low-dimensional GMM-LDCNN approach\footnote{Optimal parameters for protocol 1a were {\small $C=1024$}, {\small $B=128$}, and {\small $N_{U} = 128$}, while for protocol 1b {\small $C=512$}, {\small $B=96$}, and {\small $N_{U} = 128$}.}
outperforms local modelling of PCA-CNN features (GMM-PCA-CNN), with an average relative performance improvement of {\small $6.4\%$}.
The extended form of the proposed approach (GMM-LDCNN-xy) provides further improvements and obtains state-of-the-art results,
with an average relative performance improvement of {\small $14.9\%$} over Local GMM~\cite{anantharajah2014local}.
This demonstrates the effectiveness of local modelling over global features,
and highlights the potential to use feature learning techniques such as CNNs to learn effective local representations.

\begin{table}[!t]
\caption
  {
  \small
  Results on the Fish image dataset~\cite{anantharajah2014local}.
  The two baseline approaches, SVM-CNN and SVM-LDCNN, are presented along with the state-of-the-art local modelling approach from~\cite{anantharajah2014local} (\mbox{Local~GMM}).
  GMM-PCA-CNN uses PCA reduced features from fc-6 of the pre-trained CNN~\cite{krizhevsky2012imagenet}.
  The proposed GMM-LDCNN method uses \mbox{LDCNN} features in conjunction with GMMs.
  \mbox{GMM-LDCNN-xy} extends LDCNN features by adding the spatial location of each block.
  }
\label{tbl:fish_svm}
\vspace{-2ex}
\centering
\begin{tabular}{lcccc}
\toprule
\small
{\bf System} & \multicolumn{2}{c}{\bf Protocol 1a} & \multicolumn{2}{c}{\bf Protocol 1b} \\
 & {\bf Dev} & {\bf Eval} & {\bf Dev} & {\bf Eval} \\
\bottomrule                     
SVM-CNN                                    & 40.9 & 45.8 & 41.9 & 45.7 \\
SVM-LDCNN                                  & 39.2 & 44.2 & 40.3 & 43.5 \\
~\vspace{-1ex}                             &   &   &   &   \\
Local GMM~\cite{anantharajah2014local}     & 43.1 & 49.3 & 40.8 & 46.7\\
GMM-PCA-CNN                                & 45.7 & 51.5 & 44.0 & 47.2 \\  
GMM-LDCNN                                  & 51.8 & 55.5 & \textbf{46.4} & 49.5 \\ 
GMM-LDCNN-xy                               & \textbf{53.8} & \textbf{57.0} & 46.2 & \textbf{53.3} \\

\bottomrule
\end{tabular}
\vspace{-1ex}
\end{table}

\vspace{-2ex}
\subsection{Results on Food Dataset}
\vspace{-1ex}

We use the UEC FOOD-100 dataset which contains 100 Japanese food categories with more than 100 images for each category. 
Some images contain multiple classes and a bounding box is provided for each class.
Examples are shown in Fig.~\ref{fig:similar_fish_food}. 
Features are extracted from the bounding box only, so detection/localisation is not considered in this paper.

We use half of the images from each class for training and the other half for testing\footnote{We developed these protocols as insufficient details were provided to reproduce the experiments in~\cite{kawano14b}; our protocol files will be publicly available.}.
The training images are used for retraining the CNN and to learn the UBM model. 
The dimensionality for fc-6 is set to {\small $M=256$} based on initial experiments.
Initial experiments also indicated that the ISV extension to local modelling and including spatial {\small $(x, y)$} information in each feature vector
did not provide performance improvements.
As such, they were not used on this dataset.
We believe that ISV did not lead to increased performance as this is a closed-set problem\footnote{By closed set we mean that while the data differs between the training and testing sets, the classes in both sets are the same.}
with a high number of enrollment images,
resulting in less effective learning of a representation for session variation independent of the class. 
The spatial information did not help as the images are not accurately registered, consequently modelling the location of parts (such as the eggs in Fig.~\ref{fig:similar_fish_food}) is not useful.

\begin{figure}[!tb]
  \centering
  \includegraphics[width=0.85\columnwidth]{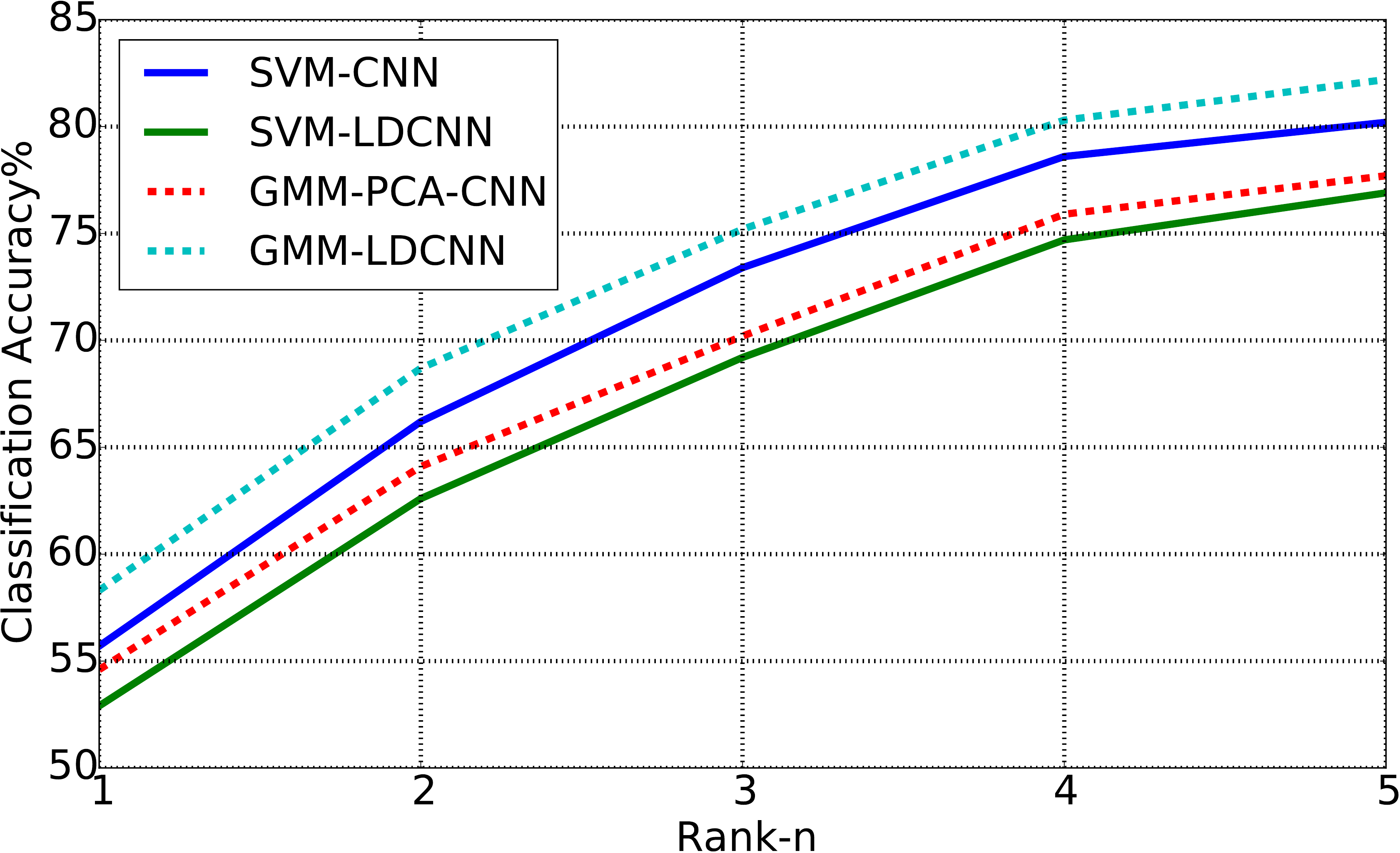}
  
  \vspace{-1ex}
  \caption
    {
    \small
    Rank-$n$ classification accuracy on the UEC FOOD-100 dataset~\cite{matsuda12}.
    }
  \label{fig:food_cnn}
  \vspace{-3ex}
\end{figure}

The results, presented in Fig.~\ref{fig:food_cnn}, show that performing local modelling using the LDCNN features (GMM-LDCNN) provides the best performance\footnote{The optimal parameters were {\small $C=512$} and {\small $B=32$}.}.
The results in Fig.~\ref{fig:food_cnn} are presented in terms of \mbox{rank-$n$} classification accuracy, where rank-$n$ refers to if the class of interest is in the $n$ best matches.
In terms of \mbox{rank-1} accuracy (identification accuracy), local modelling of the LDCNN features (GMM-LDCNN) has an accuracy of {\small $58.3\%$},
which provides a considerable relative performance improvement of {\small $9.4\%$} compared to the SVM-LDCNN approach (using LDCNN to extract a global feature) which has an accuracy of {\small $52.9\%$}.
The GMM-LDCNN approach also outperforms the SVM-CNN approach which is similar to the best single feature system presented in~\cite{kawano14b} (referred to as DCNN in their work)
and has a rank-1 accuracy of {\small $55.7\%$}.

%% file: conclusion.tex
\vspace{-2ex}
\section{Conclusion}
\label{sec:conclusion}
\vspace{-1.5ex}

In this paper we have explored the benefits of using deep convolutional neural networks (CNNs) to extract local features which are then modelled using a GMM.
Our two-step retraining procedure provides an effective way to perform dimensionality reduction and provides considerably better performance than a simple linear model such as PCA.
Comparative experiments show that considerable performance improvements can be achieved on the challenging Fish and UEC FOOD-100 datasets.

Future work will examine other ways to retrain the deep CNN.
For instance, an issue not examined in this work is the possibility of extracting thousands of local patches from each image and using these samples to retrain the entire network.